\documentclass{article}
\usepackage{spconf,amsmath,epsfig}
\usepackage{algorithm}
\usepackage{algpseudocode}
\usepackage{booktabs}
\usepackage{amsmath, amsfonts}
\usepackage{dsfont}

\usepackage{subcaption}
\usepackage{graphicx}

\let\OLDthebibliography\thebibliography
\renewcommand\thebibliography[1]{
  \OLDthebibliography{#1}
  \setlength{\parskip}{0pt}
  \setlength{\itemsep}{0pt plus 0.3ex}
}

\algnewcommand\algorithmicforeach{\textbf{for each}}
\algdef{S}[FOR]{ForEach}[1]{\algorithmicforeach\ #1\ \algorithmicdo}

\begin{document}\sloppy

\def\x{{\mathbf x}}
\def\L{{\cal L}}

\title{Multi-View Video-Based Learning: Leveraging Weak Labels for  Frame-Level Perception}

\twoauthors
 {Vijay John}
	{Guardian Robot Project\\
	RIKEN, Japan\\
	vijay.john@riken.jp}
 {Yasutomo Kawanishi}
	{Guardian Robot Project\\
	RIKEN, Japan\\
	yasutomo.kawanishi@riken.jp}

\maketitle

\begin{abstract}
For training a video-based action recognition model that accepts multi-view video, annotating frame-level labels is tedious and difficult. However, it is relatively easy to annotate sequence-level labels. This kind of coarse annotations are called as weak labels. However, training a multi-view video-based action recognition model with weak labels for frame-level perception is challenging. In this paper, we propose a novel learning framework, where the weak labels are first used to train a multi-view video-based \textit{base} model, which is subsequently used for downstream frame-level perception tasks. The \textit{base} model is trained to obtain individual latent embeddings for each view in the multi-view input. For training the model using the weak labels, we propose a novel latent loss function. We also propose a model that uses the view-specific latent embeddings for downstream frame-level action recognition and detection tasks. The proposed framework is evaluated using the MM Office dataset by comparing several baseline algorithms. The results show that the proposed \textit{base} model is effectively trained using weak labels and the latent embeddings help the downstream models improve accuracy.
\end{abstract}
\begin{keywords}
Weak Supervision, Deep Learning, Metric Learning
\end{keywords}
\section{Introduction}
\label{sec:intro}

Multi-view video-based recognition is the problem of estimating the frame-level class label from a multi-view input video. However, for training a multi-view video-based recognition model, annotating frame-level labels is a time-consuming task. On the other hand, it is far easier to annotate sequence-level labels, termed as weak labels. But, the utilization of weak labels for the frame-level perception tasks such as action recognition is challenging. Existing work in multi-view video-based learning reports state-of-the-art results~\cite{yan2021deep,li2018survey,xu2013survey}, but they do not solve the research problem of using weak labels, represented as action bags, for frame-level perception tasks. The action bag includes the labels for distinct actions within a video sequence without the start and end time of the actions. To address this limitation, in this paper, we present a novel framework using two steps to utilize the weak labels for frame-level perception tasks such as action detection and recognition. In the first step, the weak labels are used to train a multi-view video-based \textit{base} model, which is subsequently used for the downstream frame-level perception tasks. 

\begin{figure*}
\centering
\begin{subfigure}{0.45\textwidth}
    \includegraphics[width=\textwidth]{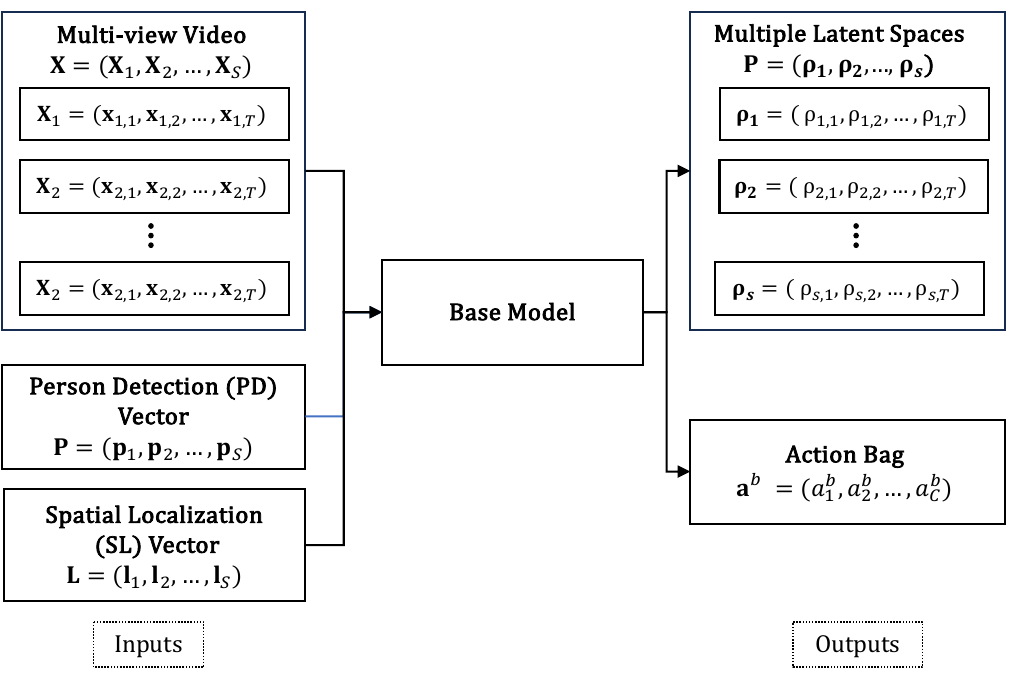}
    \caption{Base Model}
    \label{fig:first}
\end{subfigure}
\hspace{5mm}
\begin{subfigure}{0.45\textwidth}
    \includegraphics[width=\textwidth]{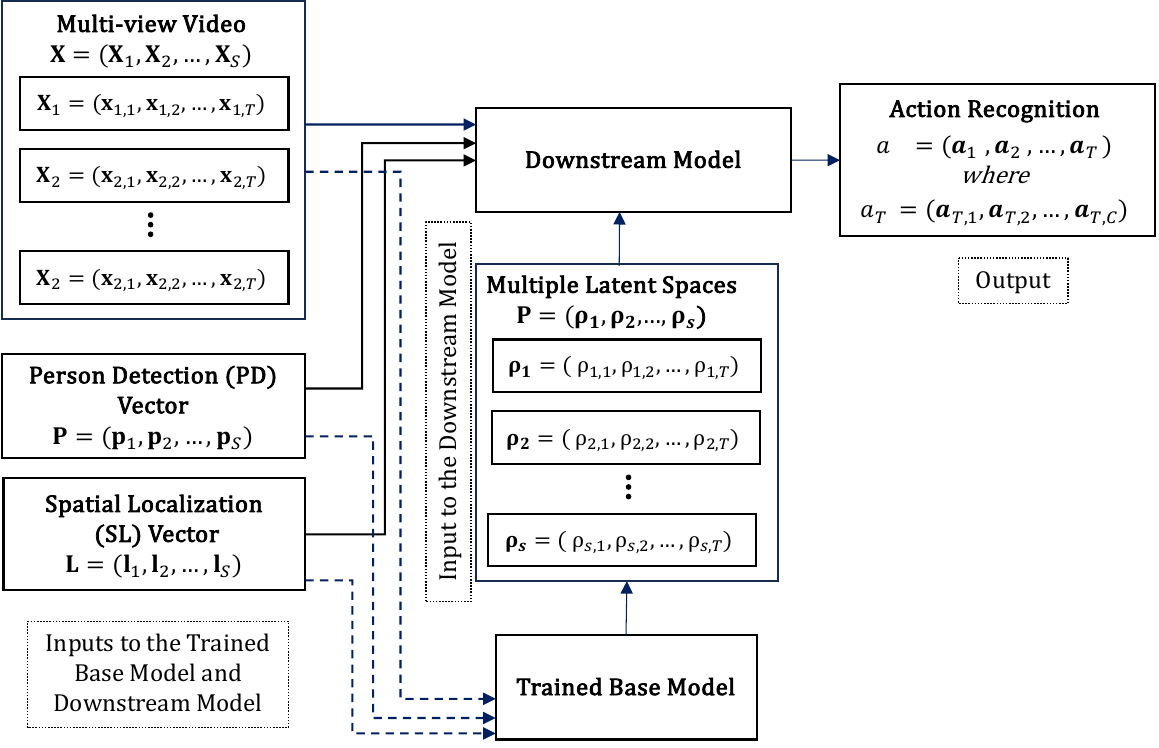}
    \caption{Downstream Model}
    \label{fig:second}
\end{subfigure}
\caption{An overview of the proposed framework.}
\label{fig:overview}
\end{figure*}

The multi-view \textit{base} model is implemented as a multi-view fusion framework using a transformer~\cite{dosovitskiy2020image} and trained using metric learning. Moreover, the outputs of the person detector are represented as additional inputs for the multi-view \textit{base} model. In the \textit{base} model, the transformer handles the long-range dependencies. Additionally, view-specific latent embeddings are learned through a novel latent loss function, termed the weak label latent loss. The weak label latent loss, based on the triplet loss, learns the latent embeddings taking into consideration the characteristics of the weak labels. The learnt latent embeddings are used for downstream tasks which handle frame-level perception. The integration of the learned latent embeddings serves as a bridge between the base and downstream models. An overview of the proposed framework is presented in Figure~\ref{fig:overview}.

The proposed framework is evaluated by using the MM Office dataset~\cite{yasuda2022multiview}. A comparative analysis with the baseline algorithms demonstrated the effectiveness of the proposed approach. This study makes the following contributions to the literature: 
\begin{itemize}
\item A novel multi-view learning framework to utilize weak labels for frame-level perception.
\item A new latent loss function, to learn the view-specific latent embedding, using weak labels.
\end{itemize}

The remainder of this paper is organized as follows: Section~\ref{sec:literature} provides a literature review of the relevant literature. The proposed framework is presented in Section~\ref{sec:proposed}. The validation of the framework and discussion of the results are presented in Section~\ref{sec:experiment}. Finally, we summarize our results and the present the directions of future research in Section~\ref{sec:conclusion}.

\section{Literature Review}
\label{sec:literature}
The literature in multi-view learning focuses on efficiently utilizing data from different sources for various perception tasks~\cite{yan2021deep,li2018survey,xu2013survey}. A detailed survey of the multi-view learning literature are presented by Yan et al.~\cite{yan2021deep}, Li et al.~\cite{li2018survey} and Xu et al.~\cite{xu2013survey}. In literature, deep learning-based multi-view learning~\cite{simonyan2014two,su2015multi,yan2021deep,wang2015deep,ma2017multi} report enhanced detection accuracy. Su et al.~\cite{su2015multi} introduce a multi-view CNN architecture for 3D shape recognition, and show that using multiple views enhances the recognition accuracy. In recent years, transformer-based multi-learning report state-of-the-art accuracy~\cite{yasuda2022multiview} for multi-view learning. Apart from deep learning, other techniques such as conditional random field (CRF)~\cite{nazerfard2010conditional} and discriminant analysis~\cite{kan2015multi} are used for multi-view learning.

While existing work has made significant contributions to multi-view learning, the problem of using weak labels for frame-level perception in multi-view learning has not been addressed in detail~\cite{tan2018incomplete,Zhao}. Amongst the literature, Yasuda et al.~\cite{yasuda2022multiview} propose a transformer framework which is trained using bags of labels for a sequence, as weak labels, for event detection. The trained transformer are then used for frame-level event detection, by extracting the outputs from the latter layers.

Compared to literature, in this paper, we propose a novel framework using two steps, where weak labels are used for frame-level perception tasks. More specifically, a multi-view \textit{base} model is first trained with weak labels to learn view-specific latent embeddings. These latent embeddings are then used for multi-view downstream frame-level perception tasks.

\section{Proposed Framework}
\label{sec:proposed}

To train the base and the downstream models, a multi-view dataset obtained from $S$ cameras is used. The multi-view synchronised data are represented as $\mathbf{X}=(\mathbf{X}_1, \ldots, \mathbf{X}_S)$, where $\mathbf{X}$ is a multi-view video data. Here, a video corresponding to the \textit{s}-th view with $T$ frames is represented as $\mathbf{X}_s=(\mathbf{x}_{s,1},\ldots, \mathbf{x}_{s,T})$. Typically, for frame-level perception tasks, the ground truth label for $T$ frames in the multi-view data is given as $\mathbf{g}=(\mathbf{g}_1,\ldots,\mathbf{g}_T)$, where $\mathbf{g}^{t}=(g_1^{t},\ldots,g_C^{t})$, where $g_C^{t} \in \{0, 1\}$ for the $C$ different classes. However, in a weakly supervised setting, the ground truth labels are not available at the frame-level. Instead, the ground truth are represented as action bags $\mathbf{g}^{b}=(g_1^{b},\ldots,g_C^{b})$, where $g_C^{b} \in \{0, 1\}$. Each element in the action bag is defined as follows,

\begin{align}
g_c^{b} =
\begin{cases}
    1 & \text{if } \exists g_{c,t} : g_{c,t} = 1 \\
    0 & \text{if } \forall g_{c,t} : g_{c,t} = 0.
\end{cases}
\end{align}

In this paper, the \textit{base} model is trained using an extensive amount of weak labels, while the downstream model is trained with the limited amount of frame-level ground truth labels.

\subsection{Person Detection}

In this work, a person detection model is used to detect and localize people in the multi-view video. This information is represented by \textit{ person detection} (PD) and \textit{spatial localization} (SL) vectors, which serve as additional inputs for the proposed framework. The PD vector provides useful information for differentiating frames from the background frames. On the other hand, the SL vector provides useful view-specific spatial information for multiview learning in certain environments, such as homes and offices. In such an environment with a fixed room layout, a person's location in an image provides an useful action cue. 

For the \textit{base} model and the downstream frame-level perception tasks, the PD and SL vectors provide additional cues for learning the latent spaces, and frame-level perception, respectively.

To generate these two vectors, the person detection model performs the detection individually on each view of the multi-view data. Subsequently, for a video sequence in the $s$-th view, the PD vector $\mathbf{p}_s=(p_{s,1},\ldots,p_{s,T}) \in \{0,1\}$ is obtained, with $1$ corresponding to every frame containing a person, and $0$ otherwise. 

\begin{figure}[!t]
 \begin{center}
\includegraphics[width=80mm]{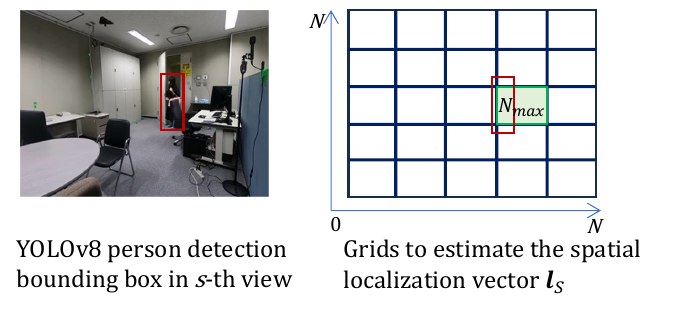}
  \end{center}
  \caption{ An overview of the estimation of the spatial localization (SL) vector.}
  \label{fig:iougrid}
\end{figure}

Next, using the bounding boxes of the detected persons, the SL vector is obtained by measuring the maximum intersection-over-union (IOU) between the bounding box of the detected person and predefined non-overlapping image grids. The non-overlapping image grids are generated for for each view. The SL vector for a sequence in the $s$-th view  is given as $\mathbf{l}_s=(l_{s,1},\ldots,l_{s,T})$, where $l_{s,T}$ is $N$-dim one-hot vector, with the \textit{n}-th grid with maximum overlap having value $1$ and the other grids having value $0$. An overview of the generation of the two vectors are presented in Fig~\ref{fig:iougrid}.

\subsection{Base Model}

\begin{figure*}
\centering
    \includegraphics[width=170mm]{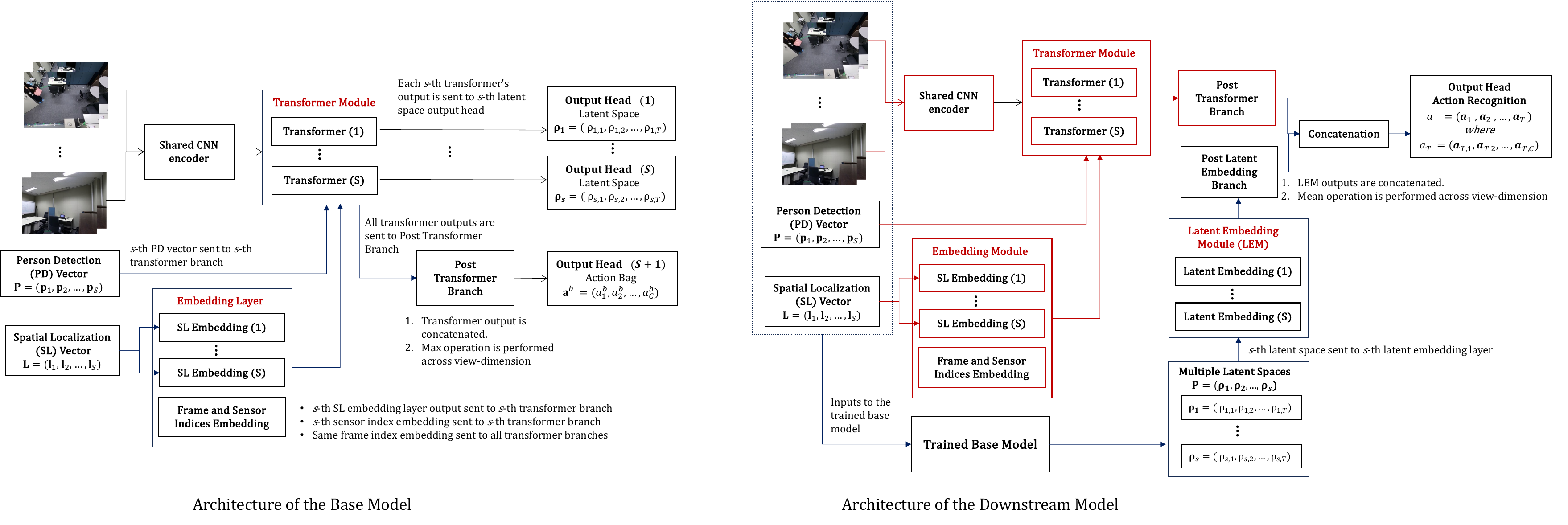}
    \caption{Architecture of the base and downstream models. In the downstream model, the modules similar to the \textit{base} model are denoted by the red rectangle.}
    \label{fig:first}
\label{fig:arch}
\end{figure*}

The \textit{base} model is implemented using CNN and the transformer~\cite{dosovitskiy2020image} and trained using metric learning. The transformer utilizes the multi-head self-attention mechanism to assign attention weights to the each view's data. Compared to the RNN and LSTM, the transformer model handles long-range dependencies in the video sequence~\cite{dosovitskiy2020image}. The weak label latent loss, based on the triplet hard loss, is designed according to the specific characteristics of the action bags $\mathbf{g}^{b}$. We next explain the different components of the \textit{base} model in detail. An overview of the \textit{base} model is presented in Fig~\ref{fig:arch}.

\textbf{Shared Encoder:} The multi-view video is sent to a shared video encoder implemented using the CNN, which extracts the features from each video frame. The extracted feature maps for the $S$ different views are given as $\boldsymbol{\Psi}=(\boldsymbol{\Psi}_1, \ldots, \boldsymbol{\Psi}_S)$. The feature map for a video sequence corresponding to \textit{s}-th is given as $\boldsymbol{\Psi}_s=(\psi_{s,1},\ldots,\psi_{s,T})$. 

\textbf{Embedding Module: } The \textit{spatial localization} (SL) and the \textit{person detection} (PD) vectors represent the additional inputs for the \textit{base} models. Each view-specific SL vector is embedded separately using a view-specific embedding layer. The embedded vector is then added to the corresponding view's feature map. The PD vector containing binary values is directly added to the feature map. 

Since the transformer does not discern the video frame indices or the multi-view camera indices, learnable embeddings of \textit{d} dimension are generated for the frame indices and the camera index. Since all the sequences in the dataset have a fixed number of frames across the different views, the \textit{same} learnt frame embeddings are added to the feature maps across the different views. In case of the camera indices, the view-specific camera embedding are added to the corresponding view's feature maps. Following the addition of the different embedding and encoding vectors, the view-specific feature map is represented as $\boldsymbol{\Psi}_s^f$.

\textbf{Transformers Module:} There are $S$ different multi-head attention branches, each utilizing $H$ heads to compute the self-attention for the corresponding view-specific feature map, $\boldsymbol{\Psi}_s^f$. Each self-attention branch, identifies the most important feature for latent embedding and event detection. The output of the different multi-head attention branches are given as, $\boldsymbol{\varphi}_{s}$.

\textbf{Post Transformer Branch (PTB):} The output of the different multi-head attention branch, $\boldsymbol{\varphi}_{s}$, are concatenated, and subsequently the max pooling of the transformer output along the view dimension is performed to obtain the ``max'' transformer output, $\boldsymbol{\varphi}_{max}$.

\textbf{Multi-task Output Head:} The \textit{base} model has $S+1$ output heads to simultaneously learn the view-specific latent embeddings $\boldsymbol{\rho}_{s}$ and predict the action bags. 

\subsubsection{Weak Label Latent Loss}

The weak label latent loss, based on the triplet hard loss, is used by the \textit{base} model to learn the view-specific latent embeddings despite the ambiguity associated with action bags.

To learn the latent spaces, the triplet loss builds triplets consisting of an anchor data, a positive data, and a negative data. The triplet loss function aims to minimize the distance between the anchor and the positive class data, $d(a, p)$, which is in the same class as the anchor and maximize the distance between the anchor and the negative class data, $d(a, n)$, which is in the different class.

In our work, a batch hard mining strategy is used to build triplets of anchor $a$, the hardest positive data $p$ and the hardest negative data $n$ for each batch. Given a batch of embedded video data obtained from an individual camera view, 
$\mathbf{B}=(\mathbf{b}_1,\ldots,\mathbf{b}_K)$ with $K$ video samples.  Here, each $k$-th video $\mathbf{b}_k=(b_{k,1},\ldots,b_{k,T})$ corresponds to the $k$-th sample with $T$ frames. Here, $b_{k,T} \in \mathbb{R}^d$. 

The batch-wise weak labels, in the form of action bags for the $K$ video samples are given as  $\mathbf{G}^{b}=(\mathbf{g}_{1}^b,\ldots,\mathbf{g}_{K}^b)$. Here, the $k$-th action bag is 
$\mathbf{g}_{k}^b=(g_{k,1}^{b},\ldots,g_{K,C}^{b})$, where $g_{K,C}^{b} \in \{0,1\}^C$ for the different classes. Details of the weak latent label loss for a given batch in a view is provided in Algorithm 1.

\begin{algorithm}[t]
\caption{Weak Label Latent Loss}
\small
\label{algo:loss}
\begin{algorithmic}[1]
    \State Batch with $K$ samples
    \State $\mathcal{L}_f=0$
    \ForEach{$c$-th class label index}
        \State From the action bags, $\mathbf{G}_{bag}$, retrieve the $c$-th class labels index alone from all the $K$ samples to obtain a \textbf{class label} vector of $K$-dim.
        \ForEach{$t$-th frame index}
            \State From the embedded data, $\mathbf{B}$
            \State Retrieve the $t$-th index data alone from all the $K$ samples $\mathbf{b}_t=(b_{t}^1,\ldots,b_{t}^K)\in \mathbb{R}^{K \times d}$
            \State Compute the triplet hard loss using the retrieved \textbf{class label} vector $\mathbf{b}_t$
        \EndFor
    \EndFor
\label{algo:loss}
\end{algorithmic}
\end{algorithm}

\subsection{Downstream Model} 

The learned latent embeddings are integrated within the downstream models for frame-level perception. The input to the downstream models are the multi-view input, the SL vector, PD vector, and the view-specific latent embeddings. The downstream model's architecture is similar to the \textit{base} model, except for the following modules, 

\textbf{Latent Embedding Module (LEM):} The transferred view-specific latent embeddings, $\boldsymbol{\rho}_{s}$, are first concatenated within the downstream model. Subsequently, a mean operation is performed on the concatenated latent embedding along the view dimension to obtain the ``mean" latent embedding $\boldsymbol{\rho}_{mean}$. The ``mean" latent embedding is then embedded using a $d$-dimensional Dense layer.

\textbf{Concatenation:} The output of the LEM module is concatenated with the ``max'' transformer output, $\boldsymbol{\varphi}_{max}$.

\textbf{Output Head:} The concatenated output is given as input to the output head performing the frame-level perception. An overview of the downstream model's architecture is presented in Fig.~\ref{fig:arch}.

\section {Experiment} 
\label{sec:experiment}

\begin{figure}[!t]
 \begin{center}
\includegraphics[width=70mm]{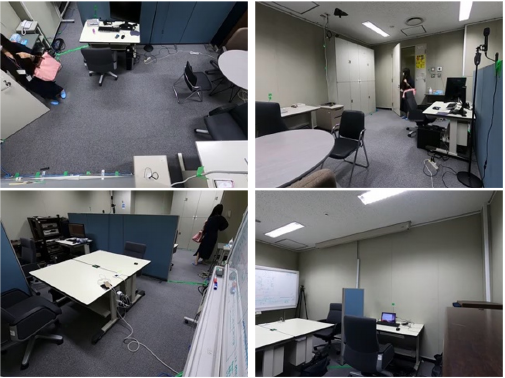}
  \end{center}
  \caption{A synchronized data from the multi-view MM office dataset~\cite{yasuda2022multiview}.}
  \label{fig:dataset}
\end{figure}

\subsection{Dataset} The proposed framework is evaluated on the public MM Office dataset~\cite{yasuda2022multiview}, which contains both weak labels and frame-level labels. The \textit{base} model is trained on $720$ multi-view synchronised videos obtained from $4$ distributed cameras with weak labels corresponding to event detection. The weak labels represented the events in a given multi-view sequence without their start/end times (Fig.~\ref{fig:dataset}). The downstream models are validated with two frame-level perception tasks, a binary action detection, and a multi-label action recognition. The downstream models are evaluated on $528$ multi-view synchronised videos with frame-level labels for action classification. Here, there are $264$ multi-view videos for training the downstream models, and $264$ multi-view videos for the testing the models. For the experiments, from each video in the multi-view data, $62$ frames are extracted at $2.5$ fps sampling rate. 

For each frame in the multi-view video, the action detection task identifies if any action is found in the frame, while the action recognition task identifies the multiple action labels in the frame.

\subsection{Implementation}
\label{ssec:algoparams}

For the person detection model, we use the pre-trained YOLOv8~\cite{yolov8_ultralytics} to generate the spatial localization and person detection vectors for each view in the multi-view input data.

\subsubsection{Base Model}
The multi-view synchronized video input is given as, $\mathbf{X}=(\mathbf{X}_1, \ldots, \mathbf{X}_4)$. Each video sequence is represented as $\mathbf{X}_1=(\mathbf{x}_{1,1},\ldots, \mathbf{x}_{1,62})$ with $62$ frames. Each \textit{t}-th frame in the \textit{s}-th video is represented as $\mathbf{x}_{s,t}\in\mathbb{R}^{(64\times64\times3)}$.

\textbf{Shared CNN Encoder:} Each image in the multi-view video is given as input to a shared CNN encoder with three Conv-2D layers with $32,64,$ and $64$ filters, $3\times3$ kernels, and ReLU activation, which are each followed by a max-pooling layer with $2\times2$ pooling. The output of the final pooling layer is given to a Dense layer with $256$ units and ReLU activation. 

\textbf{Embedding Modules:} The SL vectors are embedded to a $256$-dimensional vector using a view-specific Dense layer in the SL embedding layers. On the other hand, the frame indices and sensor indices are embedded to a $256$-dimensional vector using the Embedding layer.

\textbf{Transformers Module:} There are four transformer branches in the module for each camera view. Each transformer branch, termed as the standard transformer branch, computes the multi-head attention of $\boldsymbol{\Psi}_s^f$ using $4$ heads. The output of the $s$-th multi-head attention is added with the residual input features, $\boldsymbol{\Psi}_s^f$, using the skip connection, to generate $\boldsymbol{\Psi}_s^{f+}$. This is then projected using two dense layers with $400$ and $256$ neurons and ReLU, linear activations. The output of the second dense layer is added with the residual $\boldsymbol{\Psi}_s^{f+}$ to generate the final transformer output, $\boldsymbol{\varphi}_{s}$.

\textbf{Multi-task Output Head:} The output heads for the latent embedding, each, contains a dense layer with $256$ neurons and a linear activation function. The output of each latent embedding branch is L2 normalized and embedded using the weak label latent loss. The input to these output heads are the corresponding transformer branch's output, $\boldsymbol{\varphi}_{s}$. 

The output head for predicting the action bags consists of two Dense layers with $512$ and $12$ neurons with ReLU and sigmoid activation labels. The output of the second Dense layers provide the frame-level predictions for the different $C$ action classes. Next, based on the multiple-instance learning approach, used in weak supervised learning~\cite{kong2018sound}, the average of the frame-level predictions is performed across the frame dimension to obtain the predicted action bag. 

\subsubsection{Downstream Models}

The downstream models are similar to the \textit{base} model except the output head, and the latent inputs which are added to the output of transformer module. The output head consists of three Dense layers with $512$, $256$, and $C$ neurons, with ReLU, ReLU, and sigmoid activation functions. The $C$ corresponds to the number of classes for the frame-level perception task. The downstream models are trained with binary cross entropy function.

\subsection {Training Parameters}
The cascaded framework for both the datasets were implemented with Tensorflow 2 using NVIDIA $3090$ GPUs on an Ubuntu 20.04 desktop. The parameters for three models are selected empirically. For all the models, learning rate of $0.001$, $\beta_1$=$0.5$ and $\beta_2$=$0.99$ are used. The \textit{base} models and downstream models are trained for $100$ epochs.

\subsection{Comparative Algorithms}
For the first baseline framework, the \textit{base} model and the downstream models are based on the CNN model. Here, the architecture is similar to the proposed framework's shared video encoder and output heads. 

The second baseline framework is based on the work by Yasuda et al.~\cite{yasuda2022multiview}. Here, the \textit{base} model and the downstream models are based on the transformer~\cite{dosovitskiy2020image}. The \textit{base} model and the downstream models contain the proposed framework's shared video encoder, the embedding module, the transformer module, and output heads. For the first and second baseline frameworks, the latent space are not learnt, and the YOLOv8 outputs are also not given as inputs. 

The third baseline framework is similar to the second baseline framework. However, here, the SL and PD vectors are used as additional inputs. For the baseline frameworks, since the latent embeddings are not learnt, following its training, the pre-trained weights of the \textit{base} model are transferred and used to initialize the downstream models.

\subsection{Ablation Study}
We performed a comparative study with different variants of the proposed framework. In the first variant, termed as Ablation-\textit{A}, the SL and PD vectors are not used as inputs. In Ablation-\textit{B} and Ablation-\textit{C}, in the Post Transformer Branch (PTB), the proposed framework's max pooling operation is replaced with the sum operation and mean operation.

Finally, in Ablation-\textit{D}, a single joint latent space is learnt instead of the view-specific latent spaces. Here, in the PTB, the mean operation is performed across the view-dimension and the resulting output, $\boldsymbol{\varphi}_{mean}$, is used to learn the joint latent space. The mean operation is selected to obtain a better generalization across the different views.

\begin{table}[t!]
\small
\centering
\caption{Comparative Analysis of the Baseline Algorithms}
 \begin{tabular}{lcc}
    \toprule
    \textbf{Algorithm}  &  Action   & Action    \\
                    &     Detection & Recognition\\
    \midrule
    Proposed  & 96.7 & \textbf{63.9}  \\
    \midrule
    CNN Baseline & 93.5 & 61.1 \\
    Yasuda et al.~\cite{yasuda2022multiview} & 96.7 & 60.9 \\
    Trans. with SL and PD.~\cite{dosovitskiy2020image} & \textbf{97.7} & 61.6 \\
  \bottomrule
\end{tabular}
\label{table:baseline models}
\end{table}

\begin{table}[t!]
\small
\centering
\caption{Ablation Study of the Proposed Framework}
 \begin{tabular}{ccccccc}
    \toprule
    \textbf{Algo.}  & SL & PD & PTB  & Latent &  Action   & Action    \\
                     &  &    &  opera. & Space & Det. & Recog.\\
    \midrule
    Proposed & $\checkmark$ & $\checkmark$ & max & multiple & \textbf{96.7} & \textbf{63.9}  \\
    \midrule
    Ablation-\textit{A} & $\times$ & $\times$ & max & multiple &  96.1 & 63.3 \\
    Ablation-\textit{B} & $\checkmark$ & $\checkmark$ & sum & multiple & 95.5 & 62.3 \\
    Ablation-\textit{C} & $\checkmark$ & $\checkmark$ & mean & multiple & 94.5 & 61.4 \\
    Ablation-\textit{D} & $\checkmark$ & $\checkmark$ & mean &  single &  93.1 & 61.1 \\
  \bottomrule
\end{tabular}
\label{table:ablation study}
\end{table}

\subsection{Results and Discussion}
The experimental results show that the proposed framework is better than the baseline frameworks, especially for the difficult multi-class action recognition task. In the proposed framework, the transformer learns the long-range dependencies and assigns the attention weights to the important features. The view-specific loss efficiently learns the latent space by computing the loss from the batch-wise data for each class across the different time indices. By utilizing the \textit{base} model's latent space within the downstream models, the framework efficiently transfers the learning. Comparing the results, we can observe that the multi-class action recognition is a difficult task than the binary action detection. In case of the binary action detection task, the transformer models (proposed and baseline) report similar performance.

The advantages of the SL and the PD vectors are observed in the comparison with Ablation-\textit{A}. Next, the max operation in the PTB is shown to be better than the sum and mean operation. One possible explanation could be the multi-view dataset, where there are limited overlapping views across the cameras. Finally, the comparison with Ablation-\textit{D} learning the view-specific latent spaces is better. 

\section {Conclusion}
\label{sec:conclusion}

In conclusion, this paper introduces a novel multi-view learning framework that utilizes easily available weak labels for frame-level perception tasks. The proposed learning framework involves training a \textit{base} model with weak labels using a novel latent loss function, resulting in view-specific latent embeddings. These embeddings are subsequently used within the downstream models for frame-level perception. The proposed framework is evaluated on the MM Office dataset. A comparative analysis with baseline algorithms and an ablation study is performed. The results demonstrated the advantages of the proposed method. In future work, we will extend our method to multimodal multi-view data.

\bibliographystyle{IEEEbib}
\bibliography{icme2023template}

\end{document}